\documentclass[runningheads]{llncs}
\usepackage{array}
\usepackage{multirow}
\usepackage{hyperref}
\usepackage{amsmath}
\usepackage[T1]{fontenc}
\usepackage{xcolor}
\usepackage{marvosym}
\usepackage{orcidlink}
\usepackage{authblk}

\usepackage{graphicx,verbatim}

\begin{document}

\title{RadiomicsRetrieval: \\ A Customizable Framework for Medical Image Retrieval Using Radiomics Features
}
\titlerunning{RadiomicsRetrieval}

\begin{comment}  

\end{comment}

\author{
    Inye Na\inst{1}$^{\dagger}$ \and
    Nejung Rue\inst{2}$^{\dagger}$ \and
    Jiwon Chung\inst{1} \and
    Hyunjin Park\inst{1,2}$^{\ast}$
}

\authorrunning{I. Na and N. Rue et al.}

\institute{
    Department of Electrical and Computer Engineering, \\ Sungkyunkwan University, Suwon 16419, South Korea \\
    \and
    Department of Artificial Intelligence, \\
    Sungkyunkwan University, Suwon 16419, South Korea \\
    \email{hyunjinp@skku.edu} 
}

\renewcommand{\thefootnote}{}
\footnotetext{$^{\dagger}$ Equal contribution.}
\footnotetext{$^{\ast}$ Corresponding author.}
    
\maketitle             

\begin{abstract}
Medical image retrieval is a valuable field for supporting clinical decision-making, yet current methods primarily support 2D images and require fully annotated queries, limiting clinical flexibility. To address this, we propose \textbf{\textit{RadiomicsRetrieval}}, a 3D content-based retrieval framework bridging handcrafted radiomics descriptors with deep learning-based embeddings at the \emph{tumor level}. Unlike existing 2D approaches, \textit{RadiomicsRetrieval} fully exploits volumetric data to leverage richer spatial context in medical images. We employ a promptable segmentation model (e.g., SAM) to derive tumor-specific image embeddings, which are aligned with radiomics features extracted from the same tumor via contrastive learning. These representations are further enriched by anatomical positional embedding (APE). As a result, \textit{RadiomicsRetrieval} enables \textbf{flexible querying} based on shape, location, or partial feature sets. Extensive experiments on both lung CT and brain MRI public datasets demonstrate that radiomics features significantly enhance retrieval specificity, while APE provides global anatomical context essential for location-based searches. Notably, our framework requires only minimal user prompts (e.g., a single point), minimizing segmentation overhead and supporting diverse clinical scenarios. The capability to query using either image embeddings or selected radiomics attributes highlights its adaptability, potentially benefiting diagnosis, treatment planning, and research on large-scale medical imaging repositories. Our code is available at \url{https://github.com/nainye/RadiomicsRetrieval}.


\end{abstract}

\section{Introduction}

\begin{figure}[t]
\hypertarget{fig_intro}{\includegraphics[width=\textwidth]{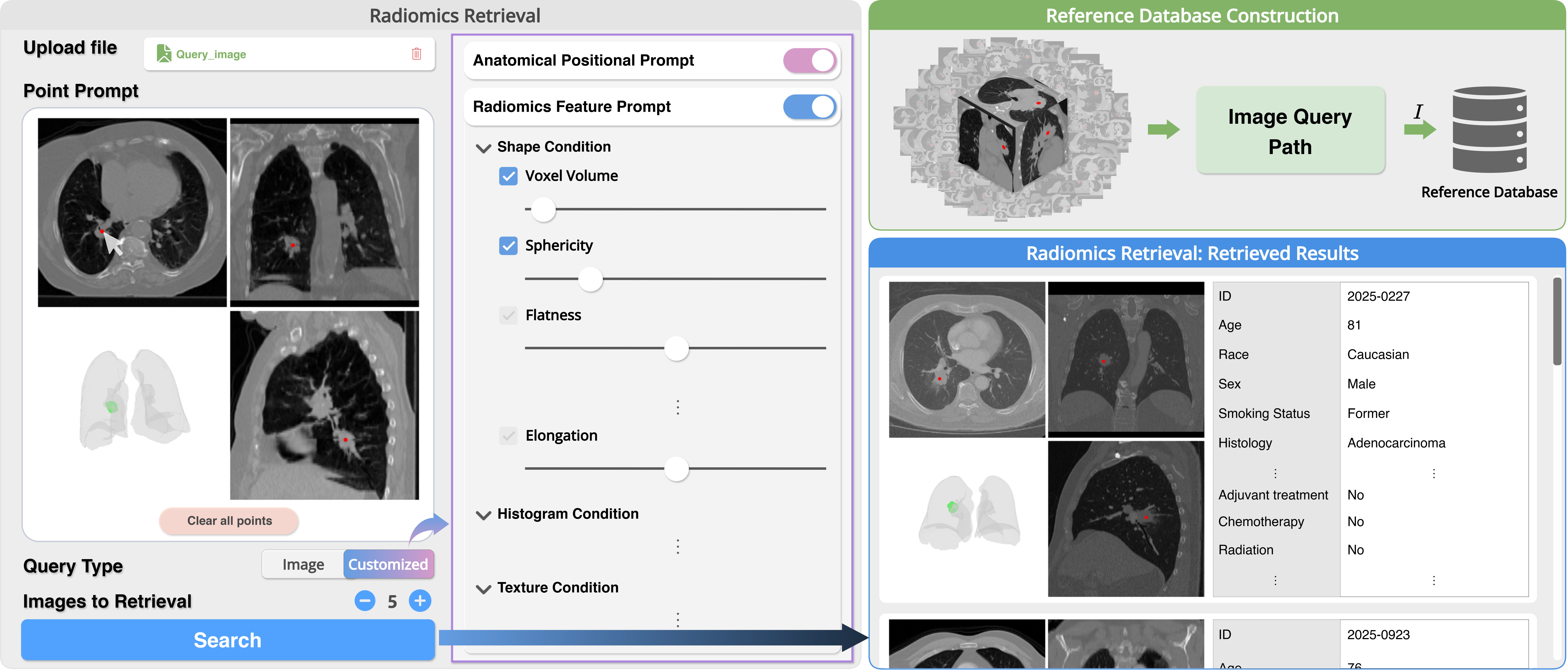}}
\caption{\textbf{Illustration of the RadiomicsRetrieval.} Users can retrieve similar tumors by providing either an image with minimal point prompts or customized conditions (e.g., anatomical location or partial radiomics features). This diverse setup supports both image-based queries and more general feature-driven searches.
}
\end{figure}

The growing volume of medical imaging has made content-based image retrieval (CBIR) a promising tool for radiologists to locate clinically relevant cases \cite{muller2004review}. While CBIR methods relying on deep learning–based encoders have surpassed those based on handcrafted features, most remain limited to 2D models that lose vital 3D anatomical context essential for accurate lesion characterization \cite{qayyum2017medical,kumar2013content,kobayashi2024sketch,abacha20233d,bottcher2025evaluation}. Current CBIR methods emphasize global image-level similarity rather than lesion-specific relevance \cite{ozturk2023content}. Although some methods accept lesion masks as queries, this is often impractical for 3D data due to the labor-intensive nature of volumetric delineation \cite{kobayashi2024sketch}. Moreover, most methods still limit queries to images, masks, or textual labels, which reduces flexibility in real-world clinical settings \cite{chen2024bimcv}.

We propose \textit{RadiomicsRetrieval}, a CBIR framework that unifies traditional radiomics with deep learning via tumor-level contrastive learning. Radiomics features, established handcrafted descriptors, capture detailed tumor shape and intensity information, but demand extensive manual segmentation \cite{na2024combined}. In contrast, deep learning methods can bypass segmentation, yet often fail to retain robust tumor specificity \cite{kumar2013content,choe2022content}. To bridge this gap, we leverage promptable segmentation models such as SAM \cite{kirillov2023segment} to derive tumor-specific image embeddings from minimal user input. Radiomics features are separately extracted from the same tumor and contrastive learning between these \textit{image embeddings} and \textit{radiomics embeddings} integrates tumor-specific attributes—shape, intensity, and spatial location—into a unified representation.

A key advantage of \textit{RadiomicsRetrieval} is \textbf{flexible querying}: users may query with an image for detailed local and global context or with radiomics features for feature-specific searches (e.g., "find highly spherical tumors in the frontal lobe"). By encoding feature names and values as text-value pairs in a language-model style, we handle varying feature sets and enable partial feature usage (Fig.~\hyperlink{fig_intro}{1}). This design broadens clinical utility and improves search efficiency, making \textit{RadiomicsRetrieval} a highly adaptable CBIR solution. Our main contributions are:

\begin{enumerate}
\item \textbf{Hybrid Radiomics–Deep Learning Approach:} Combines traditional radiomics with deep learning via tumor-level contrastive learning for enhanced clinical relevance.
\item \textbf{Flexible Query Modes:} Supports both images for detailed context and radiomics embeddings for feature-specific or partial-feature queries.
\item \textbf{Minimizing Labeling Efforts:} Uses promptable segmentation models to derive tumor-specific image embeddings from minimal prompts, reducing the burden of manual segmentation.
\end{enumerate}

\section{Method}

\begin{figure}[t]
\hypertarget{fig_method}{\includegraphics[width=\textwidth]{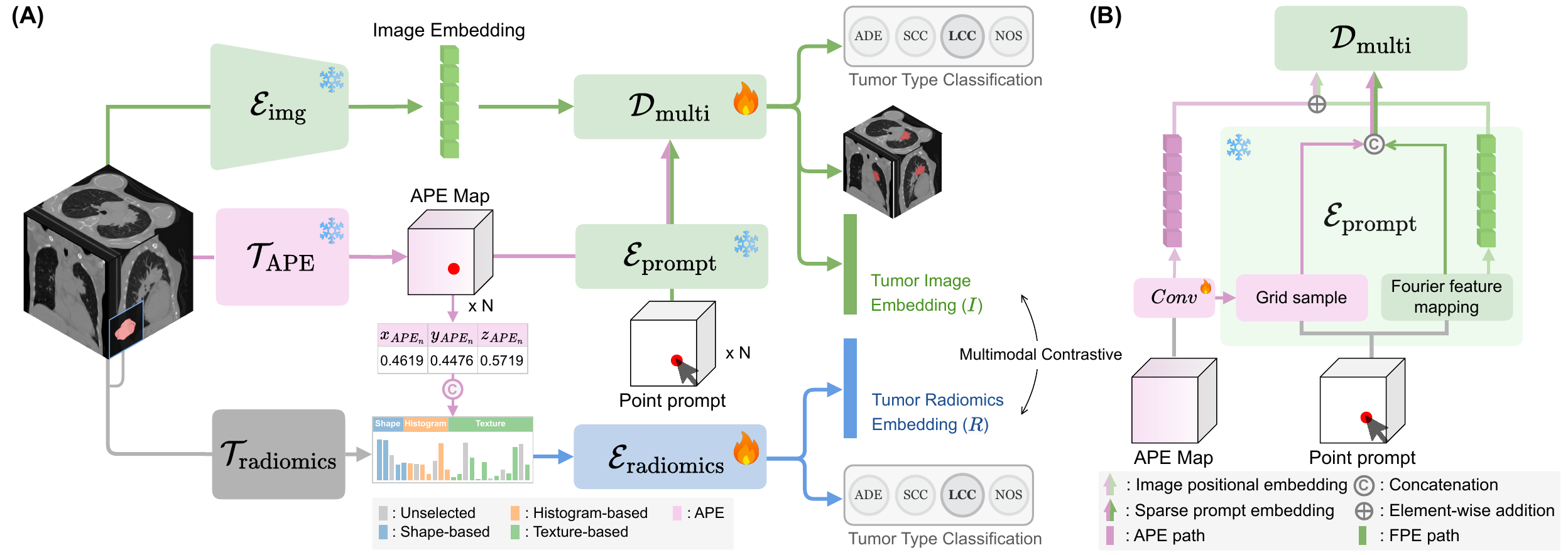}}
\caption{\textbf{(A)} \textbf{Overall architecture of \textit{RadiomicsRetrieval}}, showing parallel paths for tumor image (\textcolor[HTML]{82B366}{green}), radiomics (\textcolor[HTML]{649DE2}{blue}), and anatomical (\textcolor[HTML]{D69AC9}{purple}) embeddings. 
\textbf{(B)} Detailed view of how the anatomical positional embedding (\textcolor[HTML]{D69AC9}{APE}) is integrated with the Fourier-based positional encoding (\textcolor[HTML]{82B366}{FPE}).}
\end{figure}

\textit{RadiomicsRetrieval} operates on each tumor through two parallel paths (Fig.~\hyperlink{fig_method}{2} \textbf{(A)}): 
an \textbf{image-query path} and a \textbf{radiomics-query path}. 
The image path generates a \emph{tumor image embedding} (\(\mathit{I}\)) via a promptable segmentation model, while the radiomics path produces a \emph{tumor radiomics embedding} (\(\mathit{R}\)) from handcrafted features. 
These embeddings are aligned via \emph{tumor-level contrastive learning}, yielding a shared latent space where clinically similar lesions cluster and dissimilar lesions are pushed apart.

\subsection{Image-Query Path}
We extend SAM-Med3D~\cite{wang2023sam}, which typically outputs only a segmentation mask for the prompted region of interest (ROI), by augmenting its lightweight mask decoder with additional \emph{classification} and \emph{contrastive} tokens—denoted \(\mathcal{D}_{\mathrm{multi}}\). Given a volumetric input \((128 \times 128 \times 128)\) and a minimal prompt (e.g., a single point), the model produces a \emph{tumor image embedding} (\(\mathit{I}\)), along with auxiliary mask and classification outputs. Only during training, full segmentation masks guide the network in learning tumor-specific (ROI-level, not image-level) embeddings. However, when constructing the reference database (i.e., for retrieval), the model requires only a few point prompts per tumor, thereby reducing the need for manual ROI specification. These embeddings then populate a reference database, against which future queries are compared (Fig.~\hyperlink{fig_intro}{1}, top-right).

\subsection{Radiomics-Query Path}
A \emph{tumor radiomics embedding} (\(\mathit{R}\)) is encoded with TransTab~\cite{wang2022transtab} by tokenizing each radiomics feature as a feature name-value pair. Users can specify any subset of features without retraining, supporting flexible search conditions. We extract 72 standard radiomics features per tumor, including 14 shape-based, 18 histogram, and 40 texture features via PyRadiomics~\cite{van2017computational}. During training, we randomly sample radiomics subsets ranging from 72 to 1 feature, enhancing robustness for partial feature queries (Fig.~\hyperlink{fig_pos_neg}{3}\textbf{(B)}).

\subsection{Anatomical Positional Embedding (APE)}
We incorporate global anatomical context using a 3-channel voxel-based APE~\cite{goncharov2024anatomical}, integrating it into both image and radiomics paths (Fig.~\hyperlink{fig_method}{2}\textbf{(A)}). In the \textbf{image path}, the APE map is downsampled through convolutional layers to align with SAM's Fourier-based positional encoding (FPE)~\cite{tancik2020fourier} dimensions and then integrated element-wise (Fig.~\hyperlink{fig_method}{2}\textbf{(B)}). Additionally, sampled APE values at prompt coordinates are concatenated with sparse prompt embeddings to reinforce local anatomical context. 
In the \textbf{radiomics path}, voxel-level APE explicitly enriches tumor-intrinsic radiomics features, improving location-aware retrieval. Specifically, APE values at point-prompt locations are concatenated with radiomics subsets as input to \(\mathcal{E}_\mathrm{radiomics}\). During training, we randomly include APE and also train APE-only cases, enabling flexible querying (radiomics-only, APE-only, or combined) as illustrated in Fig.~\hyperlink{fig_pos_neg}{3}\textbf{(B)}. For lung CT (NSCLC), we use an APE model pretrained on abdominal and thoracic CT~\cite{goncharov2024anatomical}; for brain tumor MRI, we pre-trained the APE model with BraTS MRI data.

\begin{figure}[t]
\hypertarget{fig_pos_neg}{\includegraphics[width=\textwidth]{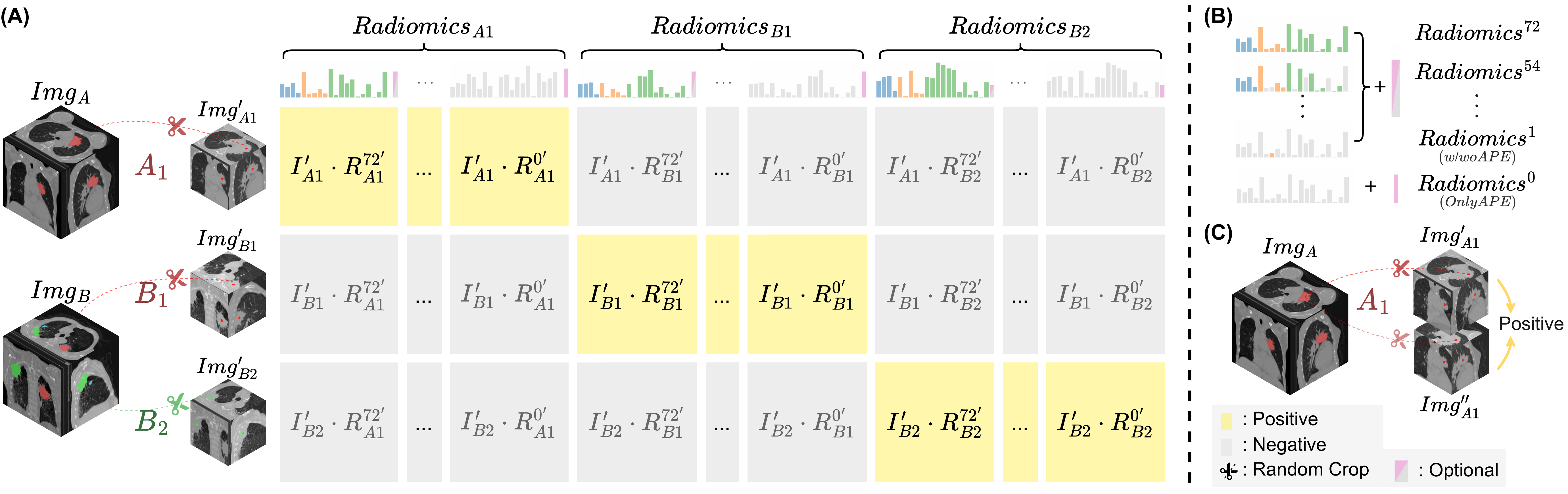}}
\caption{\textbf{Multi-positive contrastive learning.}
\textbf{(A)} Same-tumor embeddings are \emph{positives} (yellow); different tumors are \emph{negatives} (gray).
\textbf{(B)} Radiomics feature subsets.
\textbf{(C)} Paired random cropping (twice per tumor).
}
\label{fig_pos_neg}
\end{figure}

\subsubsection{Training Details}
\paragraph{Loss Functions.}
We freeze large pretrained modules (SAM and APE) and fine-tune only the lightweight decoder (\(\mathcal{D}_{\mathrm{multi}}\)) and the radiomics encoder. We jointly optimize:
(1)~\(\mathcal{L}_{\mathrm{seg}}\) (Dice + Cross-Entropy) for segmenting the prompted tumor,
(2)~\(\mathcal{L}_{\mathrm{contrast}}\) (InfoNCE-style \cite{chen2020simple}) for clustering embeddings of the same tumor and separating others, and
(3)~\(\mathcal{L}_{\mathrm{cls}}\) (Binary Cross-Entropy or Cross-Entropy) for classifying tumor type in both the image and radiomics pathways.
\paragraph{Multi-positive Contrastive Setup.}
As shown in Fig.~\hyperlink{fig_pos_neg}{3}, we use a batch size of 80, containing 40 distinct tumors, each repeated with randomly chosen point prompts (1–10) and randomized crops. This produced two image embeddings (e.g., \(\mathit{I}'\) and \(\mathit{I}''\)) per tumor and multiple radiomics embeddings (e.g., \(\mathit{R}^{72'} \dots~\mathit{R}^{0'}\), \( 
\mathit{R}^{72''} \dots~\mathit{R}^{0''}\)) incorporating matching APE values. Embeddings from the same tumor formed \emph{positives}, while embeddings from other tumors served as \emph{negatives}, enabling multi-positive contrastive learning.

\subsubsection{Retrieval Inference.}
At inference time, a user may provide (1)~an image with minimal point prompts, yielding a \emph{tumor image embedding} via the image path, or (2)~a set (or subset) of radiomics features, APE, or both, yielding a \emph{tumor radiomics embedding}. The resulting query embedding is then compared against the reference database using cosine similarity in the shared latent space, allowing both precise image-based searches and flexible feature-driven queries (e.g., all lesions with a diameter exceeding 30\,mm in the left frontal lobe).

\section{Dataset}

\subsubsection{Brain Tumor MRI.}
We used the Adult Glioma (GLI) \cite{baid2021rsna,menze2014multimodal,bakas2017advancing} and Meningioma (MEN) \cite{labella2023asnr} subsets from the BraTS 2023 Challenge. Only the tumor core, comprising enhancing tumor and necrotic areas, was considered. Patients were split (8:1:1, stratified by tumor type) into 1440 (GLI 724, MEN 716) tumors for training, 176 (GLI 89, MEN 87) for validation, and 179 (GLI 90, MEN 89) for testing. For retrieval evaluation, ANTsPy's non-linear registration aligned anatomical masks from the ICBM T1 atlas \cite{fonov2009unbiased,fonov2011unbiased,collins1999animal+}, yielding precise left/right brain parcellations.

\subsubsection{Lung Tumor CT.}
We combined non-small cell lung cancer (NSCLC) data from multiple sources (NSCLC-Radiomics \cite{aerts2019data}, NSCLC-Radiomics-Interobserver1 \cite{wee2019data}, RIDER-LungCT-Seg \cite{wee2020data}, NSCLC Radiogenomics \cite{bakr2017data}, and LUNG-PET-CT-Dx \cite{li2020large}), focusing only on NSCLC cases. The dataset was split (7:1:2) at the patient level, stratified by subtype (adenocarcinoma [ADC], squamous cell carcinoma [SCC], large cell carcinoma [LCC], not otherwise specified [NOS]). This yielded 672 (SCC 168, LCC 85, ADC 302, NOS 48, Unlabeled 69) tumors for training, 100 (SCC 24, LCC 14, ADC 45, NOS 7, Unlabeled 10) for validation, and 192 (SCC 65, LCC 35, ADC 74, NOS 18) for testing. Unlabeled tumors were ignored in the classification loss. We used TotalSegmentator \cite{wasserthal2023totalsegmentator} to obtain anatomical masks, including five lung lobes (upper/lower on each side plus the right middle lobe).

\section{Results}

We evaluated retrieval performance using the test set as queries and the training set as the reference database. Specifically, we constructed the reference database using image embeddings derived from a single tumor-center prompt per training image. To examine how radiomics features and APE influenced retrieval, we compared five training settings. We started with a baseline that used only an image path (\textit{unimodal-contrastive}) with FPE (\texttt{<Image,FPE>}), and then either kept or replaced FPE with APE or added a radiomics path (\textit{multimodal-contrastive}) as follows:

\begin{itemize}
    \item\texttt{<Image,APE>}: Replaced FPE with APE.
    \item\texttt{<Image,Radiomics,FPE>}: Kept FPE and added a radiomics path.
    \item\texttt{<Image,Radiomics,APE>}: Replaced FPE with APE and added a radiomics path.
    \item \texttt{<Image,Radiomics,FPE+APE>}: Combined both FPE and APE alongside the radiomics path.
\end{itemize}

\subsection{Evaluation of Global Context}

\begin{table}[t]
\centering
\caption{\textbf{Location-Based Evaluation.} \textit{P@k} indicates the mean precision@k, i.e., the fraction of top-$k$ retrieved samples that lie in the same region as the query. \textbf{Bold} denotes the best and \underline{underline} is the second best in a given category.}
\label{tab:locationBased}
\begin{tabular}{clc|cc|cc}
\hline
\multirow{2}{*}{Query type} & \multirow{2}{*}{Training modality} & \multirow{2}{*}{Contrastive} & \multicolumn{2}{c|}{BraTS} & \multicolumn{2}{c}{NSCLC} \\
 & & & P@5 & P@10 & P@5 & P@10 \\
\hline
\multirow{5}{*}{\shortstack{Image \\ \& \\ 1 tumor \\ point}}
    & \texttt{<Image,FPE>}               & unimodal & 0.8358 & 0.8028 & \textbf{0.9531} & \textbf{0.9167} \\
    & \texttt{<Image,APE>}               & unimodal & 0.7899 & 0.7598 & \underline{0.9271} & 0.8766 \\
    & \texttt{<Image,Radiomics,FPE>}     & multimodal  & 0.6458 & 0.6514 & 0.6927 & 0.6807 \\
    & \texttt{<Image,Radiomics,APE>}     & multimodal  & \textbf{0.9117} & \textbf{0.9145} & 0.9115 & \underline{0.9000} \\
    & \texttt{<Image,Radiomics,FPE+APE>} & multimodal  & \underline{0.9084} & \underline{0.8933} & 0.9062 & 0.8984 \\
\hline
\multirow{2}{*}{\shortstack{1 point \\ APE}}
    & \texttt{<Image,Radiomics,APE>}     & multimodal  & \textbf{0.8905} & \textbf{0.9000} & \underline{0.9000} & \textbf{0.8891} \\
    & \texttt{<Image,Radiomics,FPE+APE>} & multimodal  & \underline{0.8603} & \underline{0.8615} & \textbf{0.9010} & \underline{0.8880} \\
\hline
\end{tabular}
\end{table}

Table~\ref{tab:locationBased} reports location-based retrieval performance by measuring the fraction of retrieved tumors in the same region as the query. The \emph{Image} \& \emph{1 tumor point} query type used an image with a single tumor center prompt and the resulting image embedding for retrieval. Among \textit{unimodal-contrastive} models, \texttt{<Image,FPE>} surpassed \texttt{<Image,APE>}, indicating that FPE captured finer spatial details than the coarser, “average-patient”–style APE. Although APE alone lacked high-frequency precision, it was crucial in the \textit{radiomics-query path}, as radiomics features solely describe tumor-internal properties. The model learned global location features by integrating APE into the radiomics path during multimodal contrastive training with image embeddings.

Turning to \textit{multimodal-contrastive} approaches, those that used APE achieved the highest performance, showing that global anatomical cues improved location-specific matching. In contrast, \texttt{<Image,Radiomics,FPE>} focused on tumor-\linebreak internal features and was less effective for anatomical retrieval. Finally, even a \texttt{1 point APE} query localized tumors accurately, underscoring the flexibility of our method for purely anatomical searches.

\subsection{Evaluation of Local Context}

\begin{table}[t]
\centering
\caption{\textbf{Radiomics-Based Evaluation.} \textit{Top-$k$} is the average correlation between the query’s radiomics features and those of the top-$k$ retrieved samples. \textit{Upper Bound} selects the top-$k$ by directly comparing radiomics features in the reference set, while \textit{Random Baseline} randomly selects samples from the reference set. \textbf{Bold} denotes the best and \underline{underline} is the second best in a given category.}
\label{tab:RadiomicsBased}
\begin{tabular}{clc|cc|cc}
\hline
\multirow{2}{*}{Query type} & \multirow{2}{*}{Training modality} & \multirow{2}{*}{Contrastive} & \multicolumn{2}{c|}{BraTS} & \multicolumn{2}{c}{NSCLC} \\
 & & & Top-5 & Top-10 & Top-5 & Top-10 \\
\hline
\multirow{5}{*}{\shortstack{Image \\ \& \\ 1 tumor \\ point}}
    & \texttt{<Image,FPE>}               & unimodal & 0.8582 & 0.8476 & 0.8284 & 0.8161 \\
    & \texttt{<Image,APE>}               & unimodal & 0.8495 & 0.8418 & 0.7649 & 0.7543 \\
    & \texttt{<Image,Radiomics,FPE>}     & multimodal  & \textbf{0.9197} & \textbf{0.9177} & \textbf{0.8855} & \textbf{0.8818} \\
    & \texttt{<Image,Radiomics,APE>}     & multimodal  & 0.8959 & 0.8866 & 0.8316 & 0.8159 \\
    & \texttt{<Image,Radiomics,FPE+APE>} & multimodal  & \underline{0.8987} & \underline{0.8904} & \underline{0.8403} & \underline{0.8185} \\
\hline
\multirow{3}{*}{\shortstack{Radiomics \\ features}}
    & \texttt{<Image,Radiomics,FPE>}     & multimodal  & \textbf{0.9663} & \textbf{0.9621} & \textbf{0.9448} & \textbf{0.9324} \\
    & \texttt{<Image,Radiomics,APE>}     & multimodal  & 0.9484 & 0.9426 & 0.8839 & 0.8847 \\
    & \texttt{<Image,Radiomics,FPE+APE>} & multimodal  & \underline{0.9499} & \underline{0.9457} & \underline{0.9051} & \underline{0.8956} \\
\hline
\multicolumn{3}{c|}{Upper Bound} & 0.9794 & 0.9756 & 0.9711 & 0.9648 \\
\multicolumn{3}{c|}{Random Baseline} & 0.7745 & 0.7806 & 0.6815 & 0.6790 \\
\hline
\end{tabular}
\end{table}

Table~\ref{tab:RadiomicsBased} shows correlations of tumor-internal radiomics features between queries and retrievals. \textit{Multimodal-contrastive} methods substantially outperformed\linebreak\textit{unimodal-contrastive} approaches, demonstrating the benefit of combining image and radiomics paths. Notably, when radiomics features (without APE) were used as the query, the method focused more on tumor-internal attributes, yielding even higher correlation scores than image-based queries. In particular, \texttt{<Image,\linebreak Radiomics,FPE>} achieved the highest scores, suggesting this setting prioritized local tumor details over global anatomical cues. Meanwhile, \texttt{<Image,Radiomics,\linebreak FPE+APE>} remained competitive for local context while offering stronger location-based retrieval (see Table~\ref{tab:locationBased}), suggesting a robust trade-off between capturing tumor-specific details and anatomical positioning.

\subsection{Comparisons of Retrieved Results under Different Conditions}

\begin{figure}[t]
\hypertarget{fig_result}{\includegraphics[width=\textwidth]{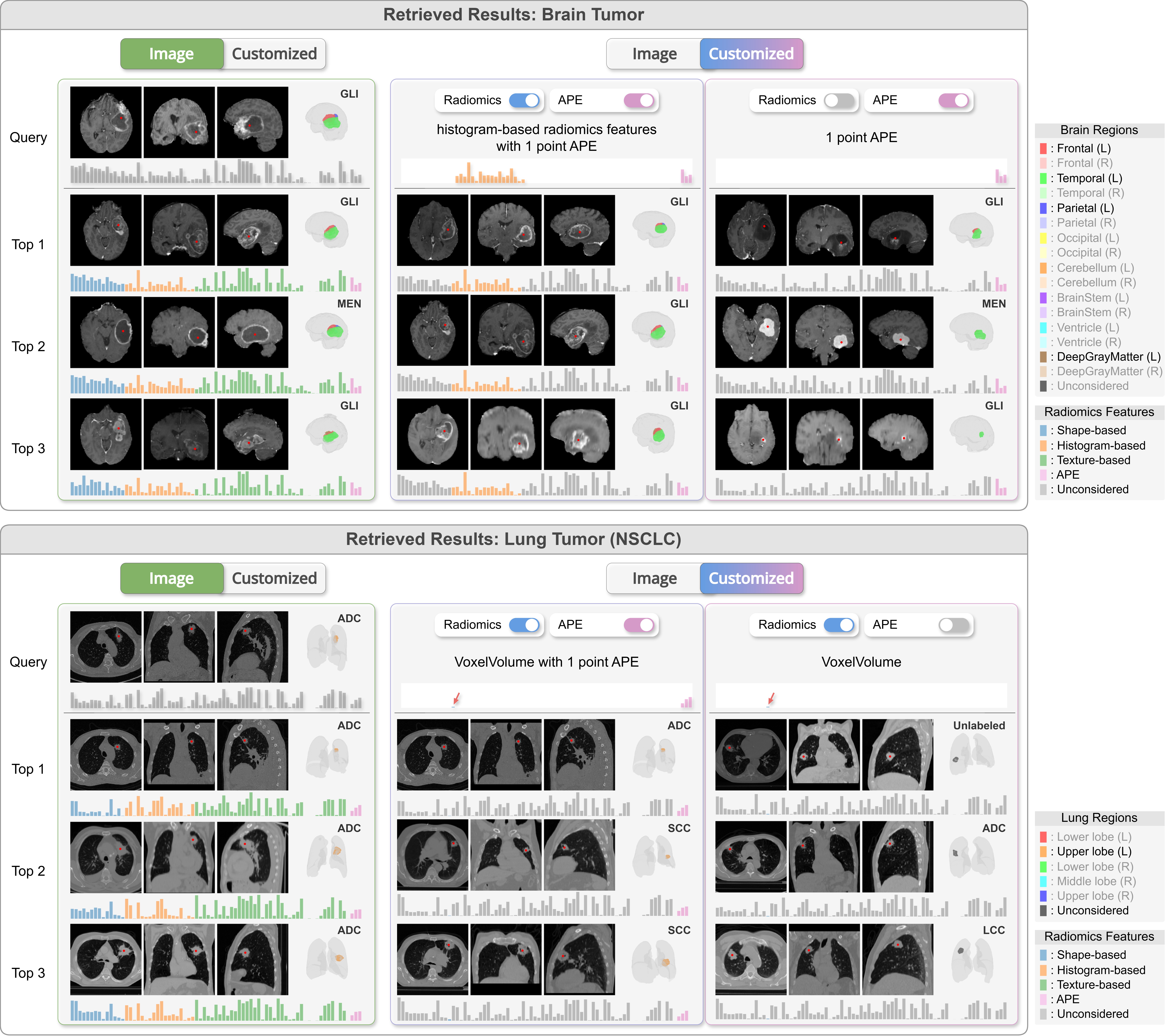}}
\caption{\textbf{Retrieved Samples Under Different Query Types.} Using the same query tumor, varying the retrieval condition (image-based, location-only, or size-only, etc.) produces distinct yet valid results, highlighting \emph{RadiomicsRetrieval}’s flexibility. The histogram under each tumor image depicts its radiomics feature values. All retrievals shown here were generated with the \texttt{<Image,Radiomics,FPE+APE>} configuration.}
\label{fig_result}
\end{figure}

Fig.~\hyperlink{fig_result}{4} illustrates how the retrieved samples vary when using different query types on the \emph{same} tumor. An \textbf{image query} with a single point prompt considers the tumor’s overall shape, size, and intensity; a \textbf{1 point APE} query focuses solely on anatomical position, potentially returning diverse tumor appearances in that region; and a \textbf{VoxelVolume feature} query emphasizes size alone, disregarding location and shape. These examples demonstrate how \emph{RadiomicsRetrieval} offers flexible, customizable searches that highlight different tumor attributes depending on the user’s clinical or research priorities.

\section{Conclusion}
We present \emph{RadiomicsRetrieval}, a flexible CBIR framework aligning image embeddings and radiomics embeddings at the tumor level via minimal prompts. By integrating APE for global context and radiomics features for tumor-specific characteristics, our method achieves robust performance across diverse query modes—including image-based, location-based, and size-based retrieval.\linebreak\emph{RadiomicsRetrieval} provides a versatile clinical tool capable of addressing various querying needs. Future directions include uncertainty quantification, systematic radiomics analyses, and query expansions.

\begin{credits}
\subsubsection{\ackname} This study was supported in part by the National Research Foundation of Korea (RS-2024-00408040), the AI Graduate School Support Program (Sungkyunkwan University) (RS-2019-II190421), the ICT Creative Consilience program (IITP-2025-RS-2020-II201821), and the Artificial Intelligence Innovation Hub program (RS-2021-II212068).

\end{credits}

\bibliographystyle{splncs04}
\bibliography{ref}

\end{document}